%% file: main.tex
\documentclass[letterpaper]{article} 
\usepackage{aaai2026}  
\usepackage{times}  
\usepackage{helvet}  
\usepackage{courier}  
\usepackage[hyphens]{url}  
\usepackage{graphicx} 
\usepackage{amsmath}
\usepackage{amsthm}
\usepackage{booktabs}
\usepackage{algorithm}
\usepackage{algorithmic}
\usepackage[switch]{lineno}
\usepackage{multirow}
\usepackage{url}
\usepackage{caption}
\usepackage{graphicx}
\usepackage{amssymb}
\usepackage{makecell}
\usepackage{booktabs}
\usepackage{subcaption}
\usepackage{dcolumn}
\usepackage{natbib}  
\usepackage{caption} 
\usepackage{algorithm}
\usepackage{algorithmic}

\usepackage{newfloat}
\usepackage{listings}
\DeclareCaptionStyle{ruled}{labelfont=normalfont,labelsep=colon,strut=off} 
\floatstyle{ruled}
\newfloat{listing}{tb}{lst}{}
\floatname{listing}{Listing}
\title{SPL-LNS: Sampling-Enhanced Large Neighborhood  Search \\ for Solving Integer Linear Programs}

\newcommand{\method}{SPL-LNS}
\author{
    Shengyu Feng,
    Zhiqing Sun,
    Yiming Yang
}
\affiliations{
    Carnegie Mellon University\\

    \{shengyuf,zhiqings,yiming\}@cs.cmu.edu
}

\usepackage{bibentry}

\begin{document}

\maketitle

\begin{abstract}
Large Neighborhood Search (LNS) is a common heuristic in combinatorial optimization that iteratively searches over a large neighborhood of the current solution for a better one.  Recently, neural network-based LNS solvers have achieved great success in solving Integer Linear Programs (ILPs)  by learning to greedily predict the locally optimal solution for the next neighborhood proposal. However, this greedy approach raises two key concerns: \textbf{(1) to what extent this greedy proposal suffers from local optima, and (2) how can we effectively improve its sample efficiency in the long run}. To address these questions, this paper first formulates LNS as a stochastic process, and then introduces SPL-LNS, a sampling-enhanced neural LNS solver that leverages locally-informed proposals to escape local optima. We also develop a novel hindsight relabeling method to efficiently train SPL-LNS on self-generated data. Experimental results demonstrate that SPL-LNS substantially surpasses prior neural LNS solvers for various ILP problems of different sizes.

\end{abstract}

\input{subtex/001introduction}
\input{subtex/005related}
\input{subtex/002preliminary}

\input{subtex/003method}
\input{subtex/004experiment}

\input{subtex/006conclusion}

\bibliography{main}


\end{document}

%% file: subtex/001introduction.tex
\section{Introduction}
Combinatorial Optimization (CO) problems present a set of fundamental challenges in computer science for decades \citep{papadimitriou1998combinatorial}. Many of those problems can be formulated as generic Integer Linear Programs (ILPs), such as logistics optimization \citep{chopra2001strategy}, 
workforce scheduling \citep{ernst2004staff}, financial portfolios \citep{milpportfolio}, compiler optimization \citep{zheng2022alpa, feng2025alternativemixedintegerlinear}, and
bioinformatic problems \citep{gusfield1997algorithms}. 
Classic ILP solvers typically conduct a tree-style search with the Branch-and-Bound (BnB) algorithm \citep{land2010automatic},  which finds the \textit{exact} solution by gradually reducing and finally
closing the gap between the primal (upper) and dual (lower) bounds of the ILPs. 
Many state-of-the-art open-source and commercial ILP solvers are of this kind, including 
SCIP \citep{achterberg2009scip}, CPLEX \citep{cplex2009v12}, and Gurobi \citep{gurobi2021gurobi}. 
However, when the problems are very large, completely closing the primal-dual gap can be intractable. Hence, solvers for large ILPs have shifted efforts
towards \textit{primal heuristics} \citep{berthold2006primal}, which are designed for finding the best possible 
solutions within a limited time window. That is, 
those heuristic primal solvers do not guarantee to find the optimal solutions, but aim to tackle large ILPs with near-optimal solutions.  Our work in this paper belongs to the category of heuristic-driven primal solvers.

Large Neighborhood Search (LNS) is a classical heuristic method for finding  
high-quality solutions much faster than pure BnB  
\citep{ahuja2002survey}. The process starts with a poor solution and iteratively revises the current solution with 
the \textit{destroy} and \textit{repair} operations. In each iteration, the system selects (or destroys) a subset of the variables, then re-optimizes (or repairs) them
 while keeping the remainders unchanged.  Finding good heuristics for the destroying and repairing operators 
 has been a central focus of the LNS solvers for solving ILPs. 
Recently,  neural network-based LNS methods have shown 
great potential because they can learn from vast amounts of data. 
For example, \citet{NEURIPS2020_e769e03a} and \citet{wu2021learning} presented a general neural LNS pipeline where the destroy operator was parameterized with a learning-based neural policy, and the repair operator was carried out by an off-the-shelf ILP solver.
 To obtain high-quality supervision, \citet{Sonnerat2021LearningAL} and \citet{huang2023searching} used a powerful expert heuristic called local branching (LB) \citep{Fischetti2003LocalB}, which guarantees the local optimal destruction policy with enough solving time. 
However, existing neural solvers are mostly trained to predict the locally optimal solution, and we find such a greedy strategy  suffering from the local optima issue (See Experiments Section)---the LNS solver may find a good solution in short time but fails to improve the objective with more searching time. In this work, our aim is to systematically investigate this local optimization problem and propose an effective strategy to improve the long-term efficiency of the sample.


To address the above challenges, we first \textbf{establish an important connection between LNS and discrete Markov Chain Monte Carlo (MCMC)}, where the probability density is defined by an energy-based model over the negative ILP objective (assuming a minimization problem). Building on this connection, we \textbf{analyze existing neural LNS proposals through the lens of locally-informed proposals} \citep{zanella2017informed}, a class of theoretically sample-efficient proposals. Motivated by this analysis, we \textbf{introduce a novel \underline{S}am\underline{PL}ing-enhanced neural \underline{LNS} method}, referred to as \method{}. \method{} is designed in the style of a locally-informed proposal and incorporates simulated annealing \citep{kirkpatrick1983SA} to effectively escape local optima. Rather than selecting the best solution in the neighborhood, \method{} samples the next candidate from a set of feasible solutions using an energy-based model. To enable efficient and informative training, we further \textbf{develop a hindsight relabeling strategy for self-supervised data generation}. Specifically, we use a trained neural network to collect LNS trajectories during the search, and retrospectively relabel the optimal destroyed variables for each step to construct high-quality supervision. These self-generated samples, combined with expert heuristic labels, are then used to train a more robust neural destroy policy within our sampling-enhanced LNS framework.

Our empirical evaluation shows that \method{} consistently outperforms the state-of-the-art neural LNS solvers and traditional heuristic methods on five ILP benchmarks, including a real-world problem. \method{} also shows strong performance in the \textit{transfer} setting where the models are trained on small instances and tested on larger ones. In a nutshell, our work advances the understanding of LNS and provides an insightful guidance on the development of neural LNS solvers.

%% file: subtex/005related.tex
\section{Related Work}

\subsection{Learning Primal Heuristic for ILPs}
The primal heuristics in ILPs aim to efficiently find high-quality feasible solutions. Diving and LNS are two main classes of primal heuristics and traditional solvers typically adopt a mixture of different variants of diving and LNS. Existing neural methods for primal heuristics mainly focus on heuristics selection \citep{ijcai2017p92, Hendel2019Adaptive, Chmiela2021LearningTS}, neural diving \citep{Nair2020SolvingMI, yoon2022confidence, han2023a, paulus2023learning} and neural LNS \citep{NEURIPS2020_e769e03a, addanki2020neural, Sonnerat2021LearningAL, wu2021learning, huang2023searching}.

LNS iteratively refines the solution by selecting a subset of decision variables (the neighborhood) to optimize at each time. Recent neural LNS methods mainly focus on the learning of neighborhood selection and leave the optimization to an off-the-shelf solver. \citet{NEURIPS2020_e769e03a} partitioned the variables into subsets via a neural model and then searched over each neighborhood sequentially. Later, \citet{wu2021learning}  and \citet{addanki2020neural} proposed more general frameworks for directly predicting the variables to optimize at each iteration.  Recently, supervised-learning-based methods achieved state-of-the-art results by learning from the expert heuristic local branching through imitation learning \citep{Sonnerat2021LearningAL} or contrastive learning \citep{huang2023searching}. However, none of the existing methods have theoretically analyze the convergence behavior of LNS. This is the first work to connect LNS with locally-informed proposals and propose a sampling-enhanced LNS method.

\subsection{Sampling for Combinatorial Optimization}
Sampling-based methods \citep{Metropolis1953EquationOS, Hastings1970MonteCS, Neal1996SamplingFM, IBA_2001} have been vastly applied in various CO problems \citep{tsp_sample, Bhattacharya2014SimulatedAA, TAVAKKOLIMOGHADDAM2007406, SECKINER200731, Chen2004MultiobjectiveVP}. However, previous methods typically suffered from a slow convergence speed compared with learning-based methods due to their inefficient proposals. The recent advances in MCMC have revitalized sampling-based methods and some of them have been successfully combined with neural models. \citet{sun2023revisiting} and \citet{feng2025alternativemixedintegerlinear} demonstrated that sampling-based methods could outperform neural CO solvers when using a locally-informed proposal. \citet{sun2023difusco} and  \citet{li2023from} successfully applied the diffusion models on CO problems, whose generative process is based on Langevin dynamics \citep{welling2011langevin}. However, the application of these sampling-based methods is greatly limited due to their incapability to generate feasible solutions directly. Instead, our proposed SPL-LNS can solve any CO problems formulated as ILPs and could be further extended to other non-linear or mixed programs when provided with an off-the-shelf solver. 

%% file: subtex/002preliminary.tex
\section{Preliminaries}
\paragraph{Integer Linear Program}
ILP is a type of discrete optimization problem whose variables are subject to integrality constraints. The general form of ILP could be expressed as
\begin{equation}
\label{eq:ilp}
\begin{aligned}
            & \min_{\mathbf{x}} \mathbf{c}^\top\mathbf{x}\\
    \text{s.t.} \ \mathbf{A}&  \mathbf{x} \leq \mathbf{b}, \ \mathbf{x}\in\mathbb{Z}^n,
\end{aligned}
\end{equation}
where $\mathbf{x}=(\mathbf{x}_1,\cdots, \mathbf{x}_n)^{\top}$ is the vector of decision variables, $\mathbf{c}\in\mathbb{R}^n$ is the vector of objective coefficients, $\mathbf{A}\in\mathbb{R}^{m\times n}$ and $\mathbf{b}\in\mathbb{R}^n$ represent the constraint coefficients. In the following context, all functions are conditioned on the input ILP and we no longer write it out explicitly. Some common metrics in measuring the solution quality includes 
\begin{enumerate}
    \item \textit{primal bound}: the objective value $\mathbf{c}^{\top}\mathbf{x}$ for the incumbent solution $\mathbf{x}$;
    \item \textit{primal gap} \citep{berthold2006primal}: the normalized difference between the primal bound and a pre-computed optimal (or best known) objective value $\mathbf{c}^{\top}\mathbf{x}^*$, defined as 
    \begin{equation}
    \begin{cases}
    \frac{|\mathbf{c}^{\top}\mathbf{x}^*-\mathbf{c}^{\top}\mathbf{x}|}{\max\{|\mathbf{c}^{\top}\mathbf{x}^*|,|\mathbf{c}^{\top}\mathbf{x}|\}}, & \text{if}  \ \mathbf{c}^{\top}\mathbf{x}^*\cdot\mathbf{c}^{\top}\mathbf{x}\geq0,\\
    1, &\text{otherwise;}
    \end{cases}
    \end{equation}

    \item \textit{primal integral}  \citep{berthold2006primal}: the integral of the  primal gap on the time range $[0,T]$.
\end{enumerate}

\paragraph{Large Neighborhood Search}
LNS is a process for iteratively improving the solution by the \textit{destroy} and \textit{repair} operators.  It starts with an initial feasible solution $\mathbf{x}^{(0)}$, typically obtained by running a traditional symbolic solver with a limited time budget. In its $t$-th iteration, the destroy operator heuristically chooses a subset of the decision variables in the current solution $\mathbf{x}^{(t)}$ and the repair operator re-optimizes the next solution $\mathbf{x}^{(t+1)}$ over the destroyed variables while keeping the remainder unchanged. Here we use $\mathbf{d}^{(t)}\in\{0,1\}^n$ to denote the destroyed set where $1$ indicates destroyed. $\eta^{(t)}=\|\mathbf{d}^{(t)}\|_1$ is the neighborhood size at $t$-th iteration. The re-optimization step is typically carried out by an off-the-shelf ILP solver that could always attain the best solution inside the preset neighborhood with enough solving time. In practice, an adaptive neighborhood size is typically used to control the difficulty of the re-optimization step
\citep{Sonnerat2021LearningAL, huang2023searching}. For example, \citet{huang2023searching} increase the neighborhood size $\eta^{(t+1)}=\min\{\gamma\eta^{(t)},\beta n\}$ when no improved solution is found, where $\gamma\geq 1$ is a constant and $\beta<1$ controls the upper bound. 


\paragraph{Local Branching}
LB \citep{Fischetti2003LocalB} is first used by \citet{Sonnerat2021LearningAL} as an expert destroy heuristic to generate the supervision for neural LNS. It formulates the optimal neighborhood selection for LNS as another ILP and searches for the next optimal solution 
$\mathbf{x}^{(t+1)}$ inside the Hamming ball with radius of $\eta^{(t)}$ centered around the current solution $\mathbf{x}^{(t)}$. LB obtains the locally optimal destroy policy that provides informative supervision for neural neighborhood selection policy.
However, LB owns the same computational complexity as the original ILP, making it unsuitable for the neighborhood selection heuristic in practice.

\paragraph{Locally-Informed Proposals}
Locally-informed proposals are developed by \citet{zanella2017informed} for efficient Markov Chain Monte Carlo (MCMC) algorithms. Consider a target distribution $p(\mathbf{x})=\exp{(-\frac{1}{\tau}E(\mathbf{x}))}/Z$ defined over $\mathbf{x}\in\mathbb{Z}^n$. Here $E(\mathbf{x})$ is an energy function, $Z=\sum_{\mathbf{x}'}\exp{(-\frac{1}{\tau}E(\mathbf{x}'))}$ stands for the normalization constant which is assumed to be intractable, and $\tau>0$ represents the temperature. Metropolis-Hastings (MH) \citep{Metropolis1953EquationOS, Hastings1970MonteCS} algorithm provides a generic MCMC framework to sample from such a distribution with rejection sampling. In each iteration, it samples $\mathbf{x}'$ from a proposal distribution $q(\mathbf{x}'|\mathbf{x}^{(t)})$, then with probability $\min\{1, \frac{p(\mathbf{x}')q(\mathbf{x}^{(t)}|\mathbf{x}')}{p(\mathbf{x}^{(t)})q(\mathbf{x}'|\mathbf{x}^{(t)})}\}$, $\mathbf{x}'$ would be accepted as $\mathbf{x}^{(t+1)}$, otherwise $\mathbf{x}^{(t+1)}=\mathbf{x}^{(t)}$. Such an iterative process forms a Markov chain $\mathbf{x}^{(0)},\mathbf{x}^{(1)},\cdots$ with stationary distribution $p(\mathbf{x})$. 

A good proposal distribution needs to balance $p(\mathbf{x}')$ and $q(\mathbf{x}^{(t)}|\mathbf{x}')$
to achieve a high acceptance rate and fast convergence. One kind of these proposals is the locally-informed proposal \citep{zanella2017informed} 
\begin{equation}
\label{eq:local_inform}
    q(\mathbf{x}'|\mathbf{x}) \propto g(p(\mathbf{x}')/p(\mathbf{x}))K(\mathbf{x'},\mathbf{x}),
\end{equation}
where $g$ is a scalar weight function (e.g., $g(y)=\sqrt{y}$ or $\frac{y}{y+1}$), and $K$ is a symmetric kernel with the same density for $K(\mathbf{x}',\mathbf{x})$ and $K(\mathbf{x},\mathbf{x}')$. The locally-informed proposals have also recently been found useful in other combinatorial optimization problems \citep{sun2023revisiting, feng2025regularized} when equipped with simulated annealing \citep{kirkpatrick1983SA}, which gradually anneals $\tau$  towards $0$, allowing the search to escape local optima and reach a low-energy state with high probability.

%% file: subtex/003method.tex
\section{Method}


\subsection{Formulating LNS as an MCMC Process}
LNS has traditionally been treated as a purely heuristic method, with limited understanding of its underlying optimization dynamics. To address this gap, we first formulate LNS as a stochastic process, drawing parallels with MCMC. Although LNS and locally-informed proposals exhibit strong similarities, no prior work has systematically explored or formalized the connection between them. We now aim to make the connection by treating the destroy and repair operators as two distinct distributions to sample from, denoted as $p_d(\cdot|\mathbf{x})$ and $p_r(\cdot|\mathbf{d}, \mathbf{x})$. Hence, each update in LNS could be treated as sampling $\mathbf{x}^{(t+1)}$ from the following distribution.
\begin{equation}
\label{eq:lns_p}
p(\mathbf{x}'|\mathbf{x}^{(t)})=\sum_{\|\mathbf{d}\|_1=\eta^{(t)}} p_r(\mathbf{x}'|\mathbf{d}, \mathbf{x}^{(t)})p_d(\mathbf{d}|\mathbf{x}^{(t)}).
\end{equation}
Consider a target distribution with the energy function
\begin{equation}
    E(\mathbf{x})=\begin{cases}
      \mathbf{c}^{\top}\mathbf{x},   &  \text{if} \ \mathbf{A}\mathbf{x}\leq \mathbf{b}, \\
      +\infty,   &  \text{otherwise}.
    \end{cases}
\end{equation}
Then LNS could be treated as an MCMC process converging to  $p(\mathbf{x})=\exp{(-E(\mathbf{x})/\tau)}/Z$ with a small $\tau$. To accelerate the convergence of this stochastic process to the target distribution, we can make the proposal distribution $p(\mathbf{x}|\mathbf{x}^{(t)})$ in the form of locally-informed proposal:
\begin{equation}
\label{eq:lns_local}
p(\mathbf{x}'|\mathbf{x}^{(t)})\propto \exp{\frac{E(\mathbf{x}^{(t)})-E(\mathbf{x}')}{2\tau}} \binom{n-d_H(\mathbf{x}', \mathbf{x}^{(t)})}{\eta^{(t)}-d_H(\mathbf{x}', \mathbf{x}^{(t)})},
\end{equation}
where $d_H(\cdot, \cdot)$ measures the Hamming distance between two inputs. The above formulation would yield the following two distributions for the destroy and repair operators:
\begin{align}   
\label{eq:d}
&p_d(\mathbf{d}|\mathbf{x}^{(t)})=\frac{Z(\mathbf{d},\mathbf{x}^{(t)})}{\sum_{\|\mathbf{d}'\|_1=\eta^{(t)}}\sum_{\mathbf{x}'\in \mathcal{S}(\mathbf{d}, \mathbf{x}^{(t)})}Z(\mathbf{d}',\mathbf{x}^{(t)})},\\
\label{eq:r}
&p_r(\mathbf{x}'|\mathbf{d},\mathbf{x}^{(t)})=\frac{\exp{(-\frac{1}{2\tau}E(\mathbf{x}'))}\mathbb{I}[\mathbf{x}'\in 
\mathcal{S}(\mathbf{d}, \mathbf{x}^{(t)})]}{Z(\mathbf{d},\mathbf{x}^{(t)})}.
\end{align}
Here, $\mathcal{S}(\mathbf{d}, \mathbf{x}^{(t)})$ represents the set of all feasible solutions to the sub-ILP induced by $\mathbf{d}$ and $\mathbf{x}^{(t)}$, and $Z(\mathbf{d},\mathbf{x}^{(t)})=\sum_{\mathbf{x}'\in \mathcal{S}(\mathbf{d},\mathbf{x}^{(t)})}\exp{(-\frac{1}{2\tau}E(\mathbf{x}')})$ is the normalizing factor. 

Examining the above equations,  we could find that existing neural LNS solvers in fact approximate the intractable target in Equation \ref{eq:d} with a mean-field distribution:
\begin{equation}
\label{eq:approx}
    p_d(\mathbf{d}|\mathbf{x}^{(t)})\approx \pi_{\theta}(\mathbf{d}|\mathbf{x}^{(t+1)})=\prod_{i=1}^n \pi_{\theta}(\mathbf{d}_i|\mathbf{x}^{(t+1)}).
\end{equation}
Note that $p_d(\mathbf{d}|\mathbf{x}^{(t+1)})$ actually represents the overall solution quality for the induced sub-ILP from $\mathbf{d}$. When $\tau$ is small, its density would concentrate on the destroyed sets that contain the locally optimal solution. Therefore, we could better approximate $p_d$ with an adjustable variance:
\begin{equation}
\label{eq:approx}
    p_d(\mathbf{d}|\mathbf{x}^{(t)})\approx \pi_{\theta}(\mathbf{d}|\mathbf{x}^{(t+1)})=\prod_{i=1}^n \pi_{\theta}(\mathbf{d}_i|\mathbf{x}^{(t+1)}),
\end{equation}
where $\pi_{\theta}(\mathbf{d}_i|\mathbf{x}^{(t+1)})$ is a Bernoulli distribution with $p=\frac{1}{1+\exp{(-\lambda^{(t)}_{i}/\sigma)}}$. Here, $(\lambda^{(t)}_{1},\cdots,\lambda^{(t)}_{n}) = f_{\theta}(\mathbf{x}^{(t+1)})$ is the output logits from the neural destroy policy $f_{\theta}$, while $\sigma>0$ controls the variance of this distribution with a similar role to $\tau$. When $\sigma\rightarrow0$, the densities of $\pi_{\theta}(\mathbf{d}|\mathbf{x}^{(t+1)})$ would also concentrate on the predicted locally optimal destroyed sets.


\subsection{Sampling-enhanced Large Neighborhood Search}
The above approximation to Equation~\ref{eq:d} has been explored in prior neural LNS solvers \citep{huang2023searching}. However, these methods can still suffer from local optima, as they do not account for the transition in Equation~\ref{eq:r}.

To address this, we propose a simple yet effective strategy to incorporate Equation~\ref{eq:r}, thereby making neural LNS a locally-informed proposal. Specifically, we leverage the feasible solutions returned by the ILP solver's repair operator: from this set, we select the top-$k$ solutions to form a candidate set and sample from it according to the energy-based model defined in Equation~\ref{eq:r}, where $k$ is a tunable hyperparameter. While seemingly simple, this approach is key to \textbf{bridging neural LNS with the framework of locally-informed proposals} and proves highly effective in escaping local optima (see the Experiments section).

Compared to the greedy update strategy, where $\mathbf{x}^{(t+1)} = \arg\max_{\mathbf{x}'} p_r(\mathbf{x}' | \mathbf{d}, \mathbf{x}^{(t)})$, our formulation introduces an additional sampling stage, transforming neural LNS into a locally-informed proposal that converges to the target distribution with theoretical efficiency guarantees. Moreover, by gradually annealing $\tau$ and $\sigma$ toward zero, SPL-LNS becomes equivalent to a simulated annealing algorithm, increasing its ability to escape local optima. The full procedure is outlined in Algorithm~\ref{alg:spllns}.

\begin{algorithm}[H]
\caption{SPL-LNS}
\label{alg:spllns}
\begin{algorithmic}[1] 
\STATE \textbf{Input}: An ILP with $(\mathbf{A},\mathbf{b},\mathbf{c})$
\STATE Find an initial feasible solution $\mathbf{x}$
\STATE Initialize $\mathbf{x}^*\gets \mathbf{x}$, $\tau$ and $\sigma$
\WHILE{stopping criterion is not met}
\STATE Sample the destroyed  variables $\mathbf{d}$ with $\pi_{\theta}$ 
\STATE Solve the sub-ILP to obtain feasible solutions $\mathcal{S}(\mathbf{d},\mathbf{x})$
\IF{$\min_{\mathbf{x}'\in\mathcal{S}(\mathbf{d},\mathbf{x})}\mathbf{c}^{\top}\mathbf{x}'<\mathbf{c}^{\top}\mathbf{x}^*$}
\STATE $\mathbf{x}^*\gets\arg\min_{\mathbf{x}'\in\mathcal{S}(\mathbf{d},\mathbf{x})}\mathbf{c}^{\top}\mathbf{x}'$
\ENDIF
\STATE Sample $\mathbf{x}'$ from $\mathcal{S}(\mathbf{d},\mathbf{x})$ with $p_r$
\IF{\texttt{Accept}($\mathbf{x}'$)}
\STATE $\mathbf{x}\gets\mathbf{x}'$
\ENDIF
\STATE Update $\tau$ and $\sigma$
\ENDWHILE
\STATE\textbf{return} $\mathbf{x}^*$ 
\end{algorithmic}
\end{algorithm} 
\begin{table*}[h!]
    \centering
    \begin{tabular}{c| c| cccc| cccc}
    \toprule
        & Real Instances &\multicolumn{4}{|c|}{Small Synthetic Instances}& \multicolumn{4}{c}{Large Synthetic Instances} \\
    \midrule
     Dataset    & Load Balancing &MVC-S & MIS-S & CA-S & SC-S & MVC-L & MIS-L & CA-L & SC-L \\
     \midrule
     \# Variables    & 61,000& 1,000 &6,000& 4,000& 4,000 & 2,000 & 12,000 & 8,000 & 8,000  \\
     \#  Constraints & 64,305 & 65,100 & 23,977 & 2,675 & 5,000 & 135,100 & 48,027 & 5,353 & 5,000 \\
     \bottomrule
    \end{tabular}
        \caption{Statistics for the problem instances in each dataset. The average number of variables and constraints are reported.}
            \label{tab:dataset_stats}
\end{table*}
\subsection{Training SPL-LNS with Hindsight Relabeling}
Although Equation~\ref{eq:approx} highlights the flexibility of reusing existing training methods, it remains unclear what the most effective data collection strategy is for training a sampling-enhanced LNS model. Since the introduction of sampling makes LNS less greedy and slows its convergence to local optima, the per-step cost of LB can be prohibitively high, making data collection inefficient.. To address this, we propose a hindsight relabeling strategy that leverages the neural LNS solver itself to generate supervision more efficiently.

Specifically, we solve an ILP from the training set with our sampling-enhanced neural LNS solver to generate samples $\mathbf{x}^{(0)}, \mathbf{x}^{(1)}, \cdots$. Then for each sample $\mathbf{x}^{(t)}$, we find the best solution within its neighborhood from the samples
\begin{equation}
    \bar{\mathbf{x}}^{(t)} = \arg\min\nolimits_{\mathbf{x}\in\{\mathbf{x}^{(0)}, \mathbf{x}^{(1)}, \cdots\}, d_H(\mathbf{x}, \mathbf{x}^{(t)})\leq \bar{\eta}^{(t)}} \mathbf{c}^{\top}\mathbf{x},
\end{equation}
where $\bar{\eta}^{(t)}$ could be different from the neighborhood size $\eta^{(t)}$ used to generate samples. We can use $\mathcal{I}_t=\{i|\bar{\mathbf{x}}^{(t)}_i\neq\mathbf{x}^{(t)}_i\}$ as the supervision to train the neural destroy policy with the binary cross-entropy loss
\begin{equation}
    L(\theta) = \sum_{t}(\sum_{i\in \mathcal{I}_t}\frac{1}{1+\exp{(-\lambda^{(t)}_i)}} + \sum_{i\notin \mathcal{I}_t}\frac{\exp{(-\lambda^{(t)}_{i})}}{1+\exp{(-\lambda^{(t)}_{i})}}).
\end{equation}
In our implementation, we first train a neural destroy policy using the samples collected by LB with the greedy update strategy. Then we use this neural LNS solver to collect more training samples with the above method. Finally, we combine the samples collected by LB and the neural solver to train another neural destroy policy for SPL-LNS. 

%% file: subtex/004experiment.tex
\section{Experiments}
\label{sec:experiment}
This section aims to empirically investigate the following research questions:

\begin{itemize}
\item \textbf{RQ1:} To what extent does the greedy neighborhood proposal suffer from the issue of local optima?
\item \textbf{RQ2:} Can the proposed SPL-LNS effectively mitigate this issue?
\item \textbf{RQ3:} How does each proposed component of SPL-LNS contibute to this effectiveness?
\end{itemize}
We begin by describing the experimental setup used to evaluate these questions.

\subsection{Experimental Setup}
\label{sec:setup}
\paragraph{Benchmark Datasets}
Since existing benchmarks are often selected to favor neural LNS solvers, we introduce a novel and more challenging dataset---\textbf{Load Balancing}---a real-world problem from the ML4CO competition \citep{gasse2022machinelearningcombinatorialoptimization}, to examine \textbf{RQ1}.

To address \textbf{RQ2} and \textbf{RQ3}, we follow prior evaluation protocols in the literature \citep{huang2023searching} and use four benchmark datasets representing diverse synthetic combinatorial optimization problems: \textbf{Minimum Vertex Covering (MVC)}, \textbf{Maximum Independent Set (MIS)}, \textbf{Combinatorial Auction (CA)}, and \textbf{Set Covering (SC)}. For each problem, we generate 200 training and 50 validation instances (all small-sized) to collect local best (LB) demonstrations. Evaluation is performed on an additional 100 small and 100 large instances, where the large instances contain twice as many variables as their small counterparts. To differentiate evaluations across instance sizes, we use the suffixes “-S” and “-L” to denote small and large instances, respectively. A summary of dataset statistics is provided in Table~\ref{tab:dataset_stats}.

We use the same procedure as in \citep{huang2023searching} to generate all instances and training data. For LB, we use $\eta^{(0)}$ as 50, 500, 200, 50 and 200  for MVC, MIS, CA, SC and Load Balancing respectively, and $\gamma$ is fixed as $1$ on all datasets. We use SCIP (version 8.0.1) to resolve ILPs formulated in LB and restrict the run-time limit to 60 minutes per iteration on each problem.

\paragraph{Baselines} We compare our methods to three non-neural baselines and two neural baselines featured by learning from LB, which include (1) BnB: the standard branch-and-bound algorithm used in SCIP (version 8.0.1), (2) RANDOM: an LNS algorithm selecting the neighborhood with uniform sampling without replacement (3) VARIABLE: an LNS algorithm with a variable neighborhood \citep{MLADENOVIC19971097} (4) LB-Relax: an LNS algorithm which selects the neighborhood with the LB-Relax heuristic \citep{huang2023local} (5) IL-LNS: a neural LNS algorithm which learns the LB heuristic through imitation learning \citep{Sonnerat2021LearningAL} and (6) CL-LNS: a recent neural LNS baseline which learns the LB heuristic via contrastive learning \citep{huang2023searching}. All of the LNS baselines adopt a greedy update strategy.
On synthetic datasets, we train all neural methods  on the demonstrations generated by LB  on small instances and evaluate them on both small and large instances in the testing set. While on Load Balancing, the neural methods are trained on the training instances with a same distribution of the testing instances. This amounts to in total $9$ testing datasets.

\paragraph{Metrics} We use the primal gap and primal integral as our evaluation metrics. To compute the primal gap, we use the best result on each instance as the optimal bound.
\paragraph{Implementation Details} The training of SPL-LNS, IL-LNS and CL-LNS is  kept the same across all datasets. We use $200$ ILP training instances and 50 validation instances to generate the training data, while each ILP instance is represented as a bipartite graph \citep{Gasse2019Exact}. A two-layer Graph Attention Network \citep{brody2022how} is used as the backbone for all neural methods. The node features from the bipartite graph are first projected into a $64$-dim embedding. We perform $2$ rounds of message passing over the bipartite graph, with $8$ attention heads and hidden size $64$.
We train each neural model for $30$ epochs, using an AdamW optimizer with learning rate $10^{-3}$ and weight decay $5\times10^{-5}$. The batch size is set as $4$. 

The training of SPL-LNS+R is basically the same as in above neural methods, except with more training data collected by relabeling. To collect training data for SPL-LNS+R, we run SPL-LNS on all training and validation instances. The LNS starts with the best solution found by LB on each instance, and all other parameters of LNS are kept the same as the ones used for inference, except the running time on each instance is restricted to $30$ minutes. For $t$-th intermediate solution, we  check along the whole trajectory to find the best solution within its $\eta^{(t)}$ neighborhood and use it to create the supervision. $\eta^{(t)}$ is set as the hamming distance between the $t$-th and $(t+1)$-th sample.

For fair comparion, we test each solver on a single CPU core of a dual AMD EPYC 7313 16-Core processor. Neural solvers are all trained and tested on a single NVIDIA RTX A6000 GPU.

\begin{figure}[hbt!]
\centering
    \includegraphics[width=.85\linewidth]{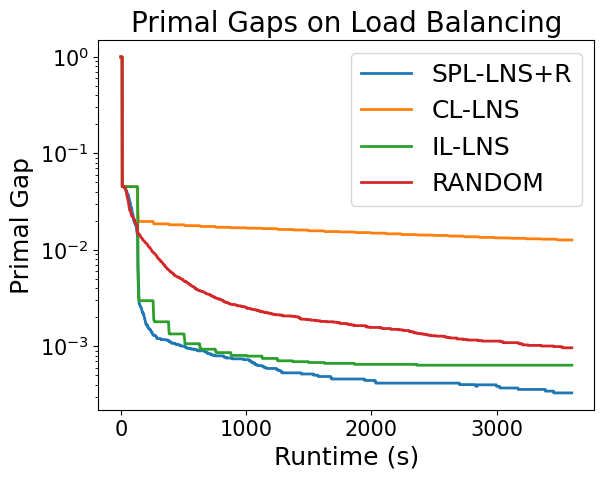}
    \caption{Comparative results on Load Balancing.} 
    \label{fig:load_balancing}
\end{figure}

\begin{figure*}[ht!]
\begin{minipage}{0.48\textwidth}
\centering
    \includegraphics[width=.8\linewidth]{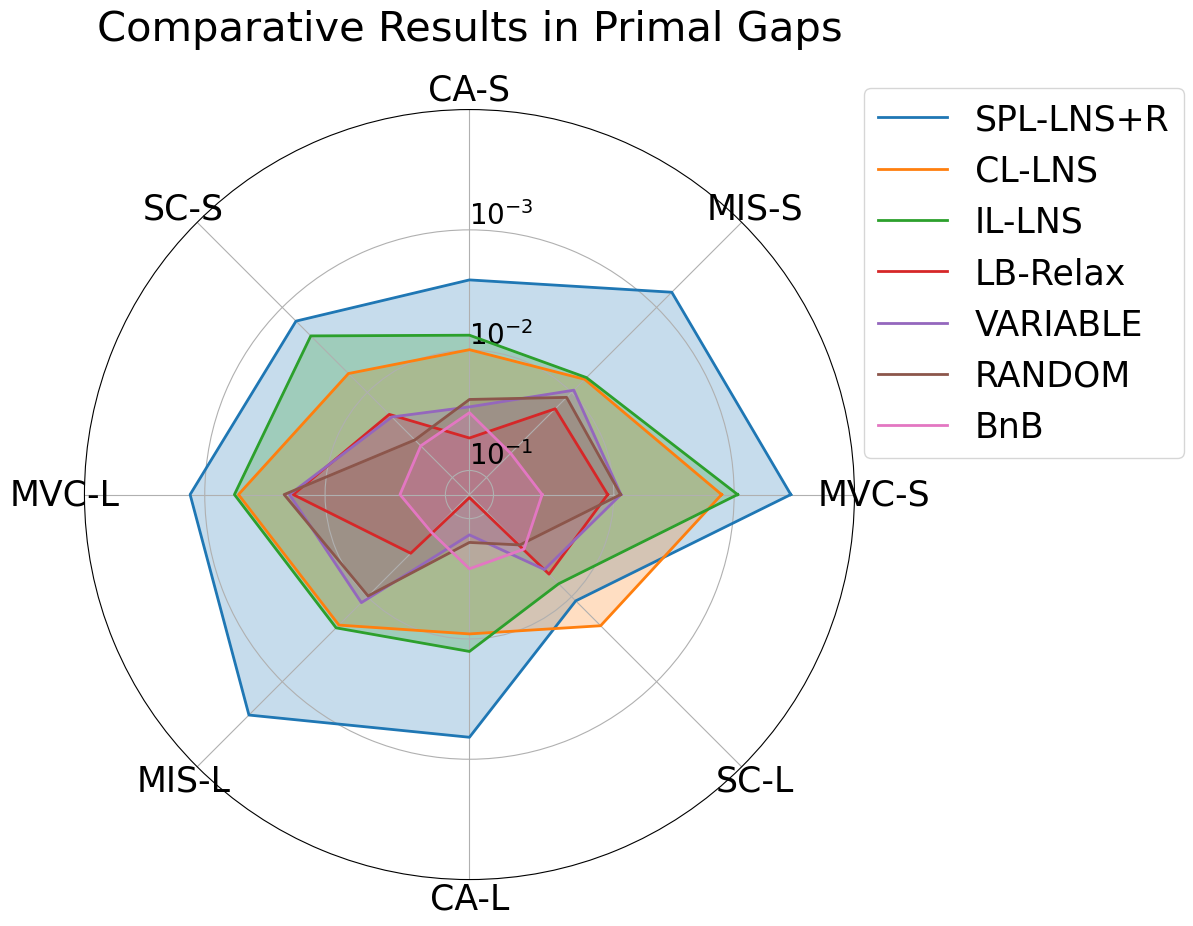}

    \caption{Comparative results in \textbf{primal gaps} at a 60-minute cutoff (lower value/outer ring is better). }  
    \label{fig:pg_spider}
\end{minipage}
\hfill
\begin{minipage}{0.48\textwidth}
\centering
    \includegraphics[width=.8\linewidth]{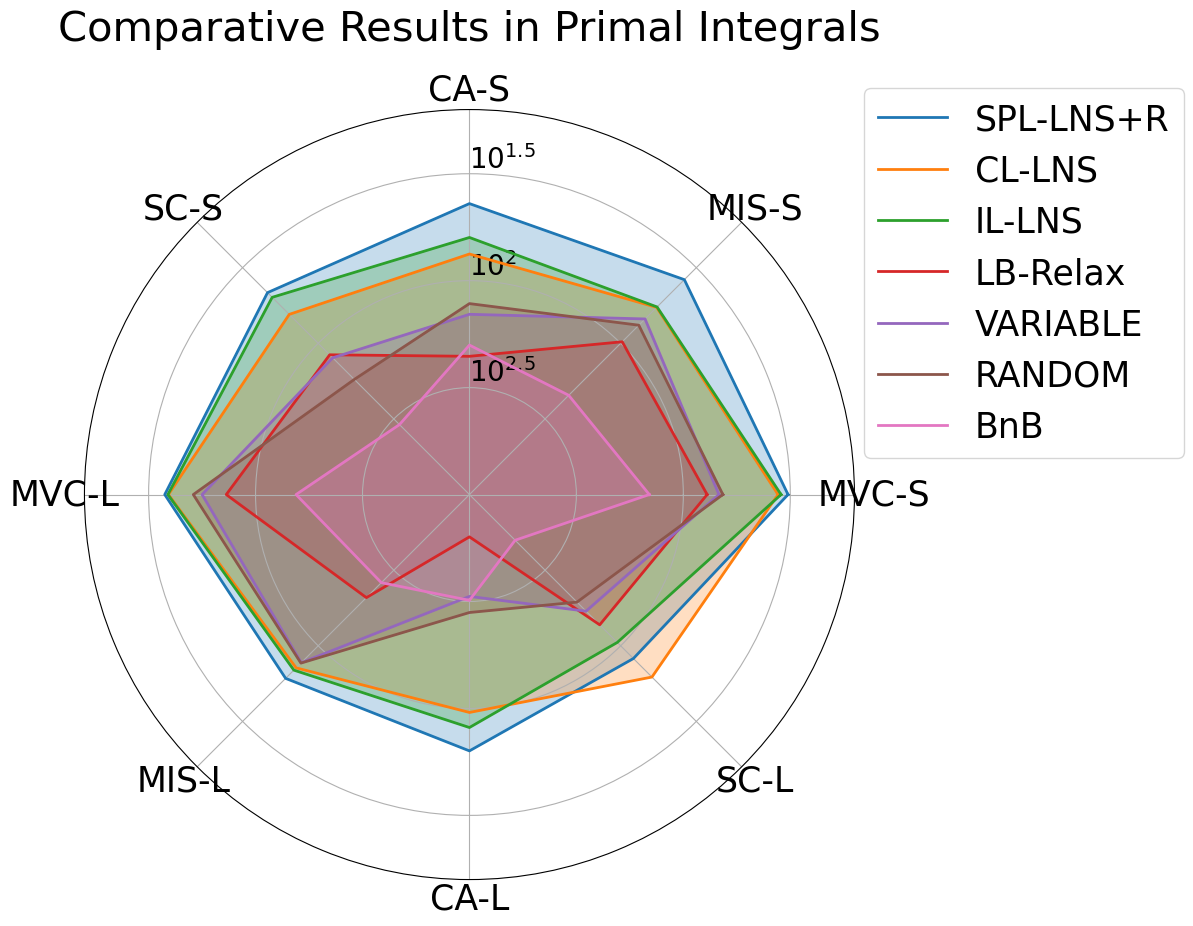}

    \caption{Comparative results in \textbf{primal integrals} at a 60-minute cutoff (lower value/outer ring is better). } 
    \label{fig:pi_spider}
\end{minipage}
\end{figure*}
\begin{figure*}[ht!]
    \centering
    \includegraphics[width=.9\linewidth]{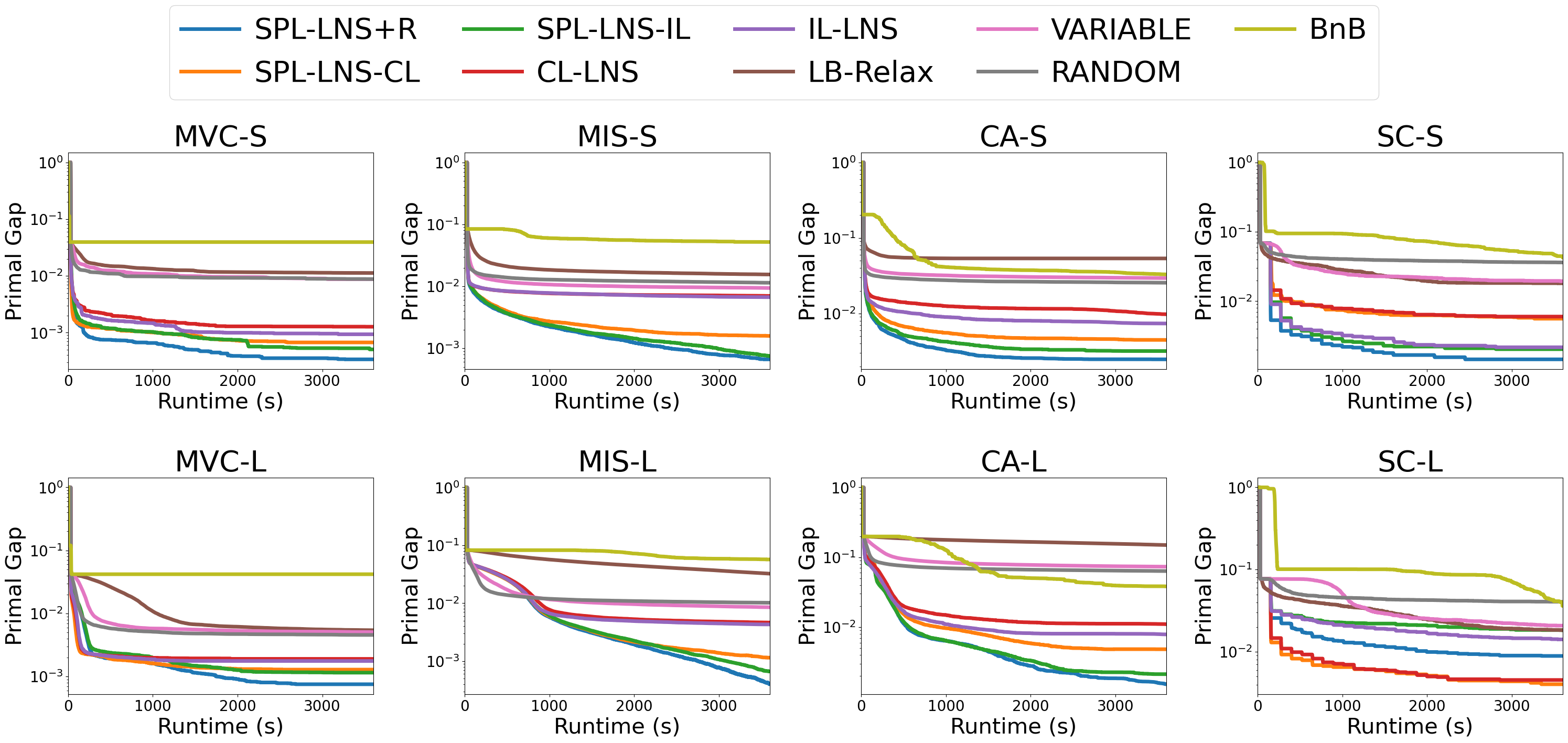}
    \caption{The plot of the primal gap (the lower is better) as a function of runtime on synthetic datasets. }
    \label{fig:pg}
\end{figure*}

\subsection{RQ1: Local Optima for Neural LNS}
\label{subsec:rq1}
To answer RQ1, we first compare neural LNS methods with the standard LNS solver with a random proposal on the Load Balancing dataset, which is adapted from a real-world application. We set the initial neighborhood size $\eta^{(0)}=500$ for the RANDOM baseline and $\eta^{(0)}=200$ for all other methods. All remaining hyperparameters are kept consistent with the previous section. The comparative results are shown in Figure~\ref{fig:load_balancing}.

The results reveal that \textbf{existing neural LNS solvers are notably affected by local optima}: their performance plateaus at a certain stage (as seen from the flat curves at the end), failing to improve further despite additional solving time. Specifically, CL-LNS stalls at a primal gap of $10^{-2}$, while IL-LNS exhibits only marginal gains over the RANDOM baseline---highlighting the limitations of greedy neural proposals in escaping local minima. In contrast, our proposed SPL-LNS+R demonstrates steady and consistent improvement over time, reducing the primal gap significantly. This trend underscores its strong potential to overcome local optima in neural LNS frameworks.


\subsection{RQ2: Effectiveness of SPL-LNS}
\label{sec:main_result}
We further examine the effectiveness of SPL-LNS with widely used synthetic benchmarks for the MVC, MIS, CA, and SC problems, following the hyperparameter settings in \citep{huang2023searching}. For VARIABLE, LB-Relax, IL-LNS, CL-LNS, and SPL-LNS, we set $\eta^{(0)}$ as $100$, $3000$, $1000$, and $150$ for MVC, MIS, CA, and SC respectively. For RANDOM, $\eta^{(0)}$ is set as $200$, $3000$, $1500$, and $200$ for MVC, MIS, CA, and SC. We fix $\gamma=1.02$ and $\beta=0.5$ in the adaptive neighborhood size for LNS-based methods across all datasets. During inference, all LNS methods start with the same feasible solution found by SCIP with a 30-second time limit. The run-time limit of SCIP on the sub-ILP $\mathbf{P}_t$ is restricted to 2 minutes. 

\begin{figure*}[ht!]
\begin{minipage}{0.24\textwidth}
\centering
\includegraphics[width=.95\linewidth]{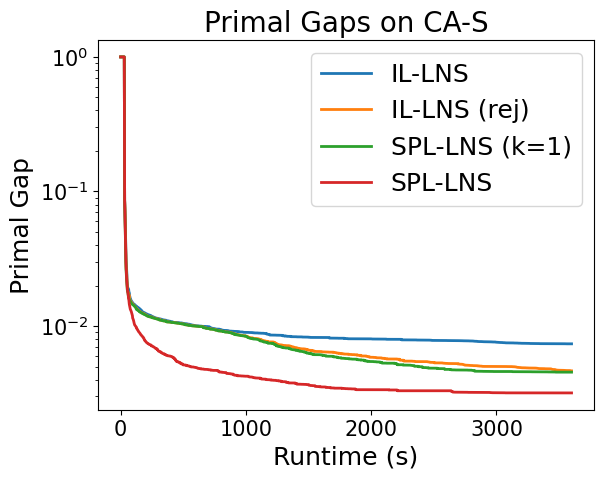}
    \caption{Comparison to other local-optima escape strategies.} 
    \label{fig:ablate_rej}
\end{minipage}
\hfill
\begin{minipage}{0.24\textwidth}
\centering
\includegraphics[width=0.95\linewidth]{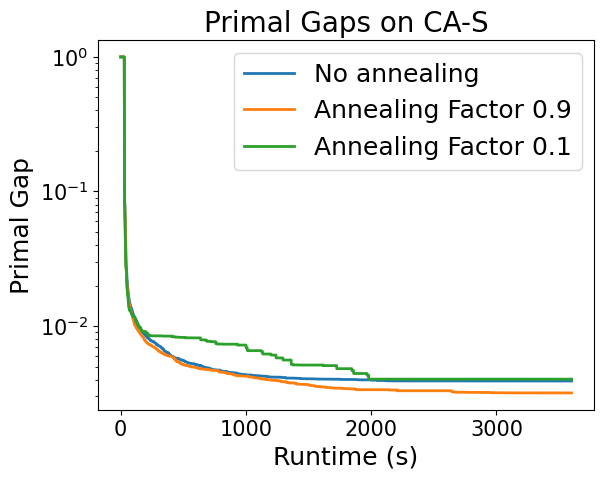}
    \caption{The ablation study on the annealing factor. } 
    \label{fig:ablate_anneal}
\end{minipage}
\hfill
\begin{minipage}{0.24\textwidth}
\centering
\includegraphics[width=0.95\linewidth]{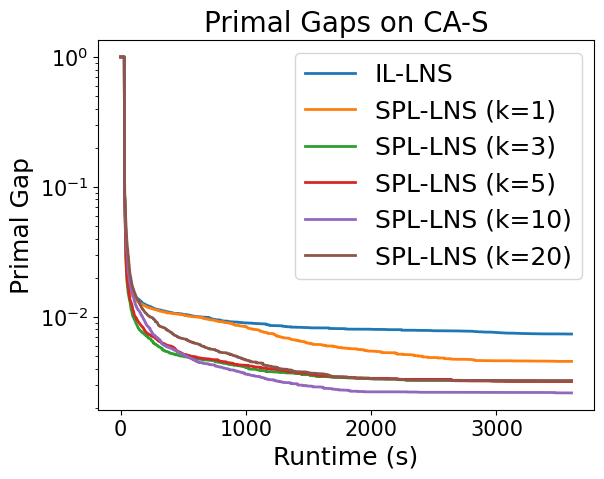}
    \caption{The ablation study on the choice of $k$.} 
    \label{fig:ablate_k}
\end{minipage}
\hfill
\begin{minipage}{0.24\textwidth}
\centering
\includegraphics[width=0.95\linewidth]{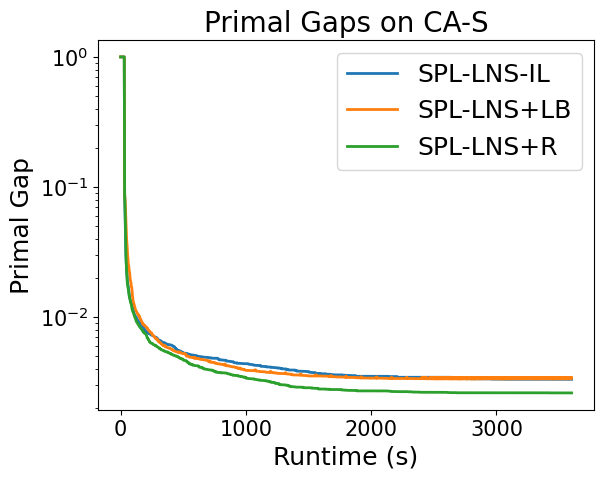}
    \caption{The ablation study on the hindsight relabeling. } 
    \label{fig:ablate_hingsight}
\end{minipage}
\end{figure*}
We follow the same schedule of $\sigma$ as in \citet{huang2023searching}, i.e., initially set as $0$ and fixed as a constant after $\eta^{(t)}=\beta n$. For the schedule of the temperature $\tau$, we let $\tau^{(0)}=|\mathbf{c}^{\top}\mathbf{x}^{(0)}|+1$ and iteratively decays it as $\tau^{(t+1)}=0.9\tau^{(t)}$. The size of the candidate set $k$ is chosen from $\{3,5\}$ based on the validation performance. We do not make careful tuning over the hyperparameters in SPL-LNS and some other choices may exhibit even better performance (e.g., see Section \ref{sec:ablation}), but SPL-LNS has already demonstrated a clear advantage over our baselines which implies its general applicability. We train SPL-LNS via both the imitation learning and contrastive learning, which are denoted as SPL-LNS-IL and SPL-LNS-CL respectively. We continue to train SPL-LNS-IL  with our hindsight relabeling method, since SPL-LNS-IL achieves an overall better performance, and we denote this method by SPL-LNS+R.

We first compare our complete sampling-enhanced LNS algorithm SPL-LNS+R with other baselines in Figure \ref{fig:pg_spider} and Figure \ref{fig:pi_spider}, where the primal gaps and primal integrals at the 60-minute cutoff are presented. It can be seen that SPL-LNS+R shows a significant advantage on all datasets except SC-L. One concern of the sampling method is the trade-off between the final performance (primal gap) and the convergence speed (primal integral), but due to the usage of the locally-informed proposal, SPL-LNS could also find a decent solution in a short time, without greatly sacrificing the primal integral.

To better understand the effects of the two proposed components, we visualize the dynamics of the primal gap on each dataset in Figure \ref{fig:pg}, including both SPL-LNS and SPL-LNS+R. In parituclar, we denote SPL-LNS-IL and SPL-LNS-CL as the method applied on top of IL-LNS and CL-LNS, respectively. It can be seen that the greatest improvement against other baselines is still from the sampling strategy used in SPL-LNS, but the hindsight relabeling strategy could further improve the performance of SPL-LNS.

\subsection{RQ3: Ablation Studies}
\label{sec:ablation}
To better understand the contribution of each component in SPL-LNS, we perform ablation studies on the key elements of the SPL-LNS algorithm. All ablation experiments are conducted on the \textbf{CA-S} dataset with \textbf{IL} training.

We compare our sampling-based strategy with two alternative methods for escaping local optima in LNS. The first applies rejection sampling with simulated annealing on the best solution found within the neighborhood (including the current one), following \citet{Hottung2019NeuralLN}. The second is a greedy update strategy that excludes the current solution, equivalent to SPL-LNS with $k=1$. We refer to these as IL-LNS (rej) and SPL-LNS ($k=1$), respectively. Results in Figure~\ref{fig:ablate_rej} show that while both alternatives improve over IL-LNS, they still underperform compared to SPL-LNS. This demonstrates the effectiveness of SPL-LNS's locally informed proposal, which balances forward likelihood $p(\mathbf{x}')$ and reverse proposal $q(\mathbf{x}^{(t)}|\mathbf{x}')$ for better exploration.

We then investigate the sampling strategy in SPL-LNS, including the annealing factor and the effect of $k$. The usage of the top-$k$ sampling makes SPL-LNS method highly robust to the choice of the annealing factor, as we show in Figure \ref{fig:ablate_anneal}. Basically, choosing any value that is reasonably large than $0$, which corresponding to no sampling, would take effect.  We continue to analyze the effect of $k$ by varying its value  from $\{1,3,5,10,20\}$ and visualize the performance change of SPL-LNS in Figure \ref{fig:ablate_k}. Remarkably, even $k=3$ brings a significant improvement over the case where $k=1$. This observation verifies the importance of sampling in escaping the local optima. The performance reaches its best when $k=10$ and drops back when $k=20$, where a clear slowdown in the convergence speed can be observed. Such a trade-off in the exploration and exploitation is actually expected, but we can see that tuning $k$ for SPL-LNS is not hard and all values greater than $1$ bring a significant improvement.

Finally, we evaluate the effectiveness of our hindsight relabeling strategy. We fix the data collection time to 10 hours and collect training data using both the standard local-best (LB) method and our hindsight relabeling approach. We then continue training SPL-LNS-IL on the LB-collected data, referring to this variant as SPL-LNS+LB. Its performance is compared to SPL-LNS+R in Figure~\ref{fig:ablate_hingsight}. The results show that SPL-LNS+LB yields minimal improvement over SPL-LNS-IL due to the low efficiency of LB data collection, whereas SPL-LNS+R demonstrates a clear performance gain.

%% file: subtex/006conclusion.tex
\section{Conclusion \& Limitation}
In this paper, we highlight the insufficient exploration issue in existing neural ILP solvers, which often leads them to become trapped in local optima. To address this challenge, we introduce SPL-LNS, a novel sampling-enhanced neural LNS solver that demonstrates strong performance across a range of ILP problems. Our contributions include a theoretical analysis of the local optima phenomenon in LNS, an innovative connection between LNS and locally-informed proposals, and the integration of this proposal to mitigate local trapping. We also develop an efficient training strategy tailored to this enhanced solver. SPL-LNS serves as a general framework that can be readily applied to a variety of neural LNS solvers, with potential applicability to broader problem domains that also suffer from local optima.

While SPL-LNS shows consistent improvements and broad applicability, it introduces several design choices within the LNS framework that warrant further investigation. A robust LNS algorithm should strike a balance between quickly finding good solutions and effectively refining them over time. In this work, we follow prior approaches in adopting a relatively greedy destroy operator, while applying our annealing strategy only to the repair phase. Future work could explore alternative combinations of these strategies for potentially greater gains. Moreover, developing a generic annealing mechanism suitable for neural solvers across heterogeneous ILP tasks remains an open and promising direction for future research.